\documentclass[letterpaper, 10 pt, conference]{ieeeconf}  

\IEEEoverridecommandlockouts                              

\usepackage{stfloats}
\usepackage{graphicx}
\usepackage{subfig}
\usepackage{booktabs}
\usepackage{cite}
\usepackage{amsmath} 
\usepackage{algorithm}
\usepackage{algpseudocode}
\usepackage{multirow}

\title{\LARGE \bf
AVM-SLAM: Semantic Visual SLAM with Multi-Sensor Fusion in a Bird’s Eye View for Automated Valet Parking
}

\author{Ye Li$^{1,2}$, Wenchao Yang$^{1,2}$, Dekun Lin$^{1,2}$, Qianlei Wang$^{1,2}$, Zhe Cui$^{1,2}$, and Xiaolin Qin$^{1,2}$
\thanks{$^{1}$Chengdu Institute of Computer Applications, Chinese Academy of Sciences, Chengdu, Sichuan 610041, China.
        {\tt\small yale.li.cn@gmail.com cuizhe@casit.com.cn, qinxl2001@126.com}}%
\thanks{$^{2}$University of Chinese Academy of Sciences, Beijing 100049, China}
}

\begin{document}

\maketitle
\thispagestyle{empty}
\pagestyle{empty}

\begin{abstract}

Accurate localization in challenging garage environments—marked by poor lighting, sparse textures, repetitive structures, dynamic scenes, and the absence of GPS—is crucial for automated valet parking (AVP) tasks. Addressing these challenges, our research introduces AVM-SLAM, a cutting-edge semantic visual SLAM architecture with multi-sensor fusion in a bird's eye view (BEV). This novel framework synergizes the capabilities of four fisheye cameras, wheel encoders, and an inertial measurement unit (IMU) to construct a robust SLAM system. Unique to our approach is the implementation of a flare removal technique within the BEV imagery, significantly enhancing road marking detection and semantic feature extraction by convolutional neural networks for superior mapping and localization. Our work also pioneers a semantic pre-qualification (SPQ) module, designed to adeptly handle the challenges posed by environments with repetitive textures, thereby enhancing loop detection and system robustness. To demonstrate the effectiveness and resilience of AVM-SLAM, we have released a specialized multi-sensor and high-resolution dataset of an underground garage, accessible at https://yale-cv.github.io/avm-slam\_dataset, encouraging further exploration and validation of our approach within similar settings.

\end{abstract}

\section{INTRODUCTION}

Automated valet parking (AVP) has recently gained momentum as a solution to alleviate parking congestion and enhance driver convenience. It is typically employed in semi-enclosed spaces with low speeds and no passengers, making it well-suited for achieving Level 4 autonomous driving. In this context, the primary technical challenge is precise mapping and localization. However, parking garages often present challenges, such as poor lighting, limited texture diversity, repetitive architectural layouts, changing environmental conditions, and the absence of GPS signals. These factors pose substantial obstacles for traditional localization methods.

\begin{figure}[t]
    \captionsetup{font=small,labelfont=bf}
    \centering
    \includegraphics[width=\linewidth]{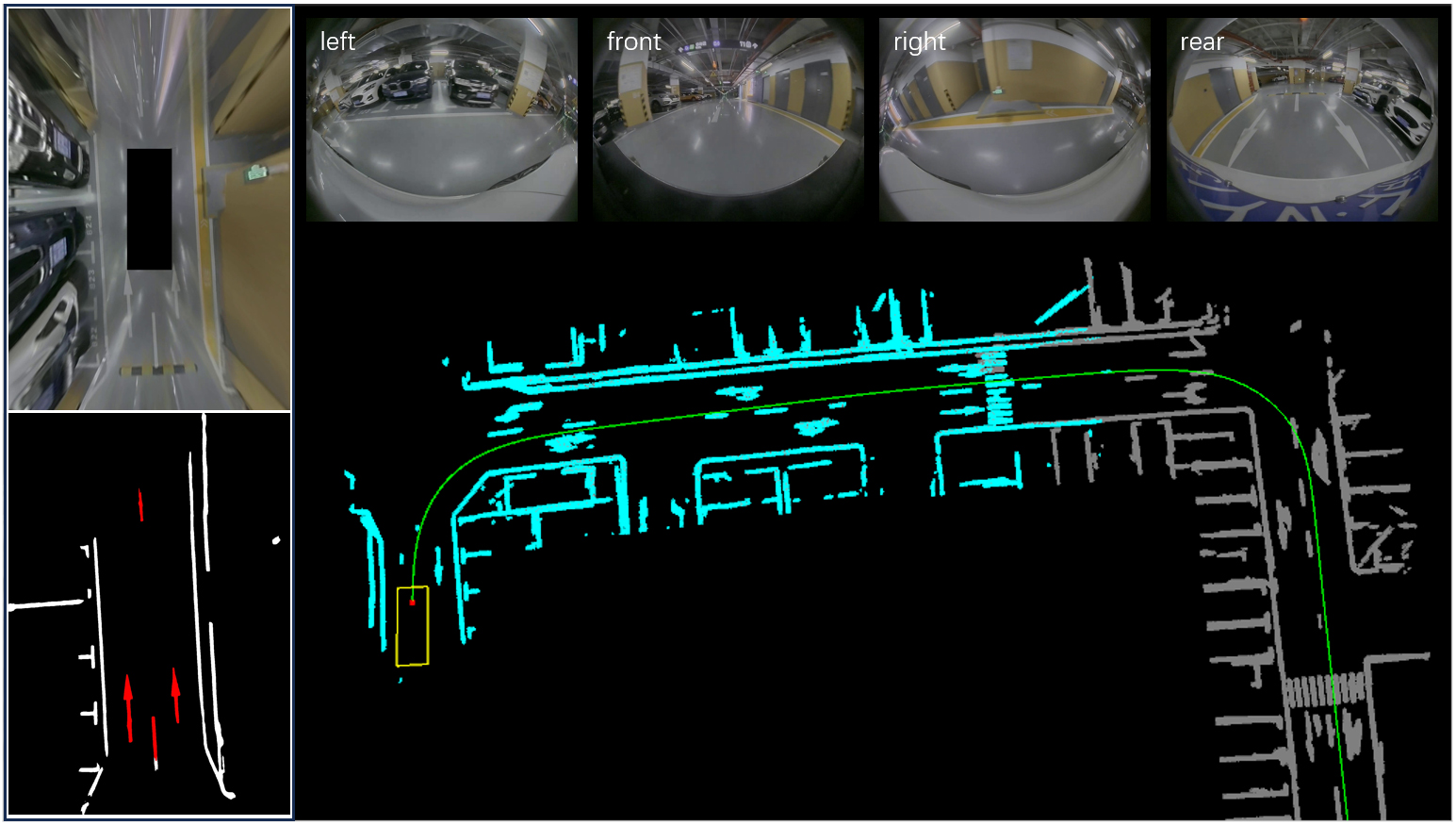}
    \caption{\small Semantic visual map of the garage built by our AVM-SLAM system. It fuses data from surround view cameras, wheel encoders and an IMU in a bird’s eye view.}
    \label{fig_system}
    \vspace{-1em}
\end{figure}

To tackle this challenge, we employ semantic attributes derived from garage road markings to build the map and localize the vehicle, as shown in Fig. \ref{fig_system}. These semantic characteristics offer enduring stability and perspective invariance, including lane lines, parking spots, zebra crossings, and indicating arrows. However, these road markers are susceptible to interference from diverging flares in bird's eye view (BEV), and for the first time, we performed flare removal on BEV images to help convolutional neural networks better extract road markers from BEV images. The BEV image is captured by surrounding cameras and generated by the around view monitor (AVM) subsystem.

In certain extreme scenarios, the extraction of semantic features might encounter difficulties. As technology has progressed, wheel encoders and inertial measurement units (IMUs) have become more affordable and nearly standard equipment in vehicle sensor arrays. Consequently, a multi-sensor hybrid fusion approach has been devised, distinct from both loosely and tightly methods, to amalgamate data from various sensors. This approach contributes to the development of the AVM-SLAM system, which stands for "a semantic visual SLAM with multi-sensor fusion in a BEV perspective."

However, the road markings in the garage, despite their persistent stability and perspective invariance, are heavily repetitive and similar, which is a disaster for those who aim to further improve the accuracy and robustness of the SLAM system through loop detection and global optimisation. Therefore, we develop a semantic pre-qualification (SPQ) mechanism that effectively prevents false matches in the complex repetitive texture environment of a garage to enhance the loop detection success rate, thereby further improving the accuracy and robustness of the system.

Furthermore, a multi-sensor and high-resolution dataset has been created and published, comprising synchronized multi-sensor data collected within an underground garage. This dataset serves to validate the effectiveness and robustness of the aforementioned methods. In summary, the primary contributions of this article are outlined as follows:

\begin{itemize}
    \item The flare removal was performed for the first time on BEV images and better semantic segmentation results for road markings were obtained.
    \item A semantic pre-qualification (SPQ) mechanism is designed to effectively prevent mismatches in environments with repetitive textures, improving the success rate of loop detection and further enhancing the system's accuracy and robustness.
    \item A multi-sensor and high-resolution dataset of typical underground garage data, contains images from four surround cameras, one synthesized BEV, measurements from four wheel encoders and an IMU. This dataset will be beneficial for further research in SLAM, particularly in the context of autonomous vehicle localization in underground garages.
\end{itemize}

\section{RELATED WORK}

\subsection{Traditional Visual SLAM and LiDAR SLAM}

Over the past decade, the field of traditional SLAM have witnessed the emergence of numerous exceptional solutions. These solutions typically employ conventional texture and 3D structural features of the environment for mapping and localization purposes. Examples include ORB-SLAM \cite{mur2015orb, mur2017orb, campos2021orb}, which relies on the ORB visual features, SVO \cite{forster2014svo} and DSO \cite{engel2017direct}, both leveraging optical flow. Additionally, Cartographer \cite{hess2016real}, LOAM \cite{zhang2014loam} and LeGO-LOAM \cite{shan2018lego} are based on 2D/3D LiDAR structural features. These approaches have demonstrated impressive performance in scenarios characterized by ample lighting, rich textures, and well-defined structures.

However, both visual and LiDAR sensors have inherent limitations. Visual sensors are sensitive to texture and can fail in scenarios with poor textures, low lighting, or overexposure. LiDAR is sensitive to structural information and can fail in structure-scarce environments. Garages, for instance, present complex scenes with poor lighting, repetitive structures, and changing scenes, where these types of methods that rely on traditional features can easily fail.

\subsection{Methods of Multi-Sensor Fusion}

In recent years, SLAM technology has shifted towards multi-sensor fusion, progressing from visual-inertial fusion to LiDAR-inertial fusion and ultimately fusing LiDAR, visual, inertial sensors, wheel encoders, and GPS data. Notable advancements include VINS-Mono \cite{qin2018vins}, VINS-Fusion \cite{qin2019general} and OpenVINS \cite{geneva2020openvins} for visual-inertial fusion, LIC-Fusion \cite{zuo2020lic} and VIL-SLAM \cite{shao2019stereo} for lidar-visual-inertial fusion, \cite{wu2017vins} and \cite{liu2019visual} for enhanced fusion with wheel encoders, VIWO \cite{lee2020visual} for sliding-window filtering to fuse multi-modal data, and \cite{niu2020accurate} for introducing wheel encoder pre-integration theory and noise propagation formula, enabling tight integration with sensor data.

By amalgamating data from multiple sensors, these approaches significantly enhance simultaneous localization and mapping robustness and accuracy. However, these methods still rely on traditional texture features and structural features, which are prone to failure in a complex environment such as a garage.

\subsection{Semantic Visual SLAM for AVP}

Semantics-enhanced visual SLAM is crucial for autonomous driving, particularly in AVP applications. Challenges in garage environments persist for both visual and LiDAR SLAM, driving the need for enduring semantic features.

Research by H. Grimmett \cite{grimmett2015integrating} and U. Schwesinger \cite{schwesinger2016automated} explores mapping and localization using surround cameras and semantic features. Z. Xiang \cite{xiang2019vilivo} extracts free space contours to simulate LiDAR points for pose tracking. J. Hu \cite{hu2019mapping} and T. Qin \cite{qin2020avp} develop a comprehensive visual SLAM system using road markings for mapping and parking facility localization. X. Shao \cite{shao2020tightly, shao2023slam} establishes tightly-coupled semantic SLAM with visual, inertial, and surround-view sensors. Z. Xiang \cite{xiang2021hybrid} utilizes hybrid edge information from bird's-eye view images to enhance semantic SLAM, while C. Zhang \cite{zhang2021avp} leverages HD vector map directories for parking lot localization. These studies offer valuable insights for future research.

In this paper, we perform flare removal for the first time on the BEV for the complex lighting conditions in the garage to better extract semantic features. Meanwhile, for the large number of repetitive textures in the garage, a SPQ mechanism is proposed to improve the success rate of loop detection. Finally, we provide a multi-sensor and high-resolution underground garage dataset and develop an AVM-SLAM system to obtain better accuracy and robustness.

\section{FRAMEWORK}

\begin{figure*}[th]
    \captionsetup{font=small,labelfont=bf}
    \centering
    \includegraphics[width=\linewidth]{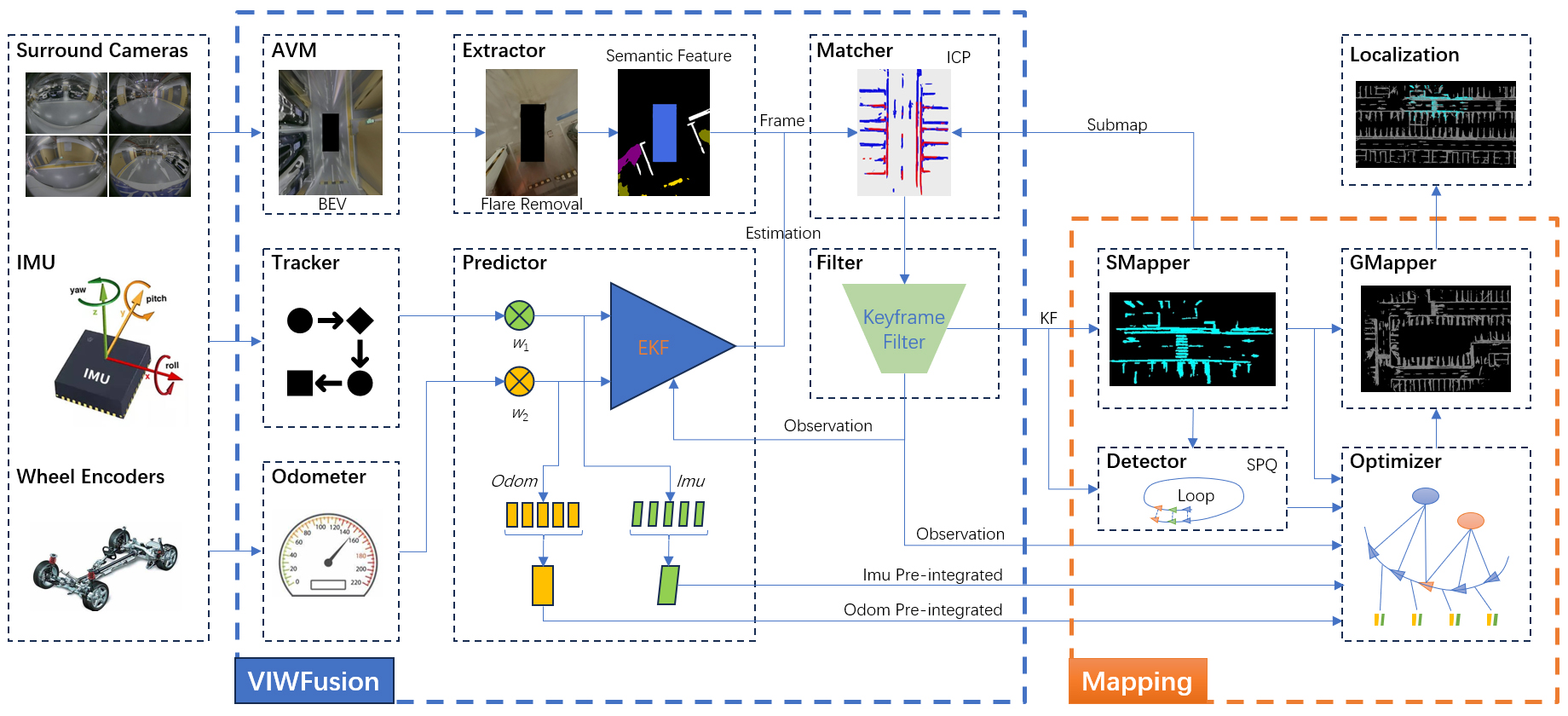}
    \caption{\small The framework of the proposed AVM-SLAM system consists of two core modules: VIWFusion and Mapping. VIWFusion is a loosely multi-sensor weighted fusion front-end, while the Mapping module serves as a tightly integrated semantic mapping back-end. $w_1$ and $w_2$ are the fusion weights for the IMU and wheel odometry, respectively.}
    \label{fig_framwork}
    \vspace{-0.5em}
\end{figure*}

This paper introduces the AVM-SLAM system, consisting of two core modules: VIWFusion and Mapping, as shown in Fig. \ref{fig_framwork}. Our design utilizes a unique multi-sensor hybrid fusion strategy, departing from traditional approaches, ensuring seamless collaboration between these modules for maximum multi-sensor fusion benefits.

VIWFusion is a loosely-coupled, multi-sensor weighted fusion front-end, encompassing the AVM subsystem, Semantic Extractor and Matcher, IMU Tracker, Wheel Odometer, Pose Predictor, and Keyframe Filter. It applies weighted fusion to data from surround cameras, wheel encoders, and IMU sensors, rooted in Extended Kalman Filter (EKF) theory, providing initial values for visual semantic matching and kinematic constraints for subsequent back-end optimization through pre-integrated (IMU and wheel) values among adjacent semantic keyframes.

The Mapping module is a tightly-coupled, semantic mapping back-end, comprising the Loop Detector, Global Optimizer, Sub Mapper and Global Mapper. We utilize semantic ICP registration for loop detection, incorporating the SPQ mechanism to streamline loop detection and minimize mismatches. Additional multi-sensor kinematic constraints, like pre-integrated values of IMU and Wheel between adjacent keyframes, expedite global optimization convergence and enhance mapping accuracy.

\section{METHOD}

\subsection{Around View Monitor} 

The AVM is crucial for generating BEV images and enhancing the SLAM system's perceptual range and robustness. It undistorts and applies inverse perspective mapping (IPM) to fisheye images from the four surrounding cameras, merging them into a comprehensive BEV image. The four surrounding cameras are strategically positioned around the vehicle, with offline calibrated intrinsic and extrinsic parameters. The virtual BEV camera is precisely located above the vehicle's center, with its optical axis aligned vertically downward. Relevant intrinsic and extrinsic parameters for this virtual camera are derived through the IPM process.

The IPM projection can be formulated as the follows:$$
\left [\begin{array}{c} u_{bev} \\ v_{bev}  \\ 1 \end{array} \right ] = H * \\
\left [\begin{array}{c} u_{undist} \\ v_{undist}  \\ 1 \end{array} \right ], \\
\eqno{(1)}
$$where $\left [\begin{array}{cc} u_{undist} & v_{undist} \end{array} \right ]$ is the pixel location in the undistortion fisheye image. $\left [\begin{array}{cc} u_{bev} & v_{bev} \end{array} \right ]$ denotes the pixel location in the bird’s eye view. $H$ represents the homography matrix from undistortion fisheye image to BEV.

\subsection{Semantic Extractor and Matcher}

\begin{itemize}
    \item Flare Removal
\end{itemize}

\begin{figure}[t]
    \captionsetup{font=small,labelfont=bf}
    \vspace{-4pt}
    \centering
        \subfloat[origin]{\includegraphics[width=0.31\linewidth]{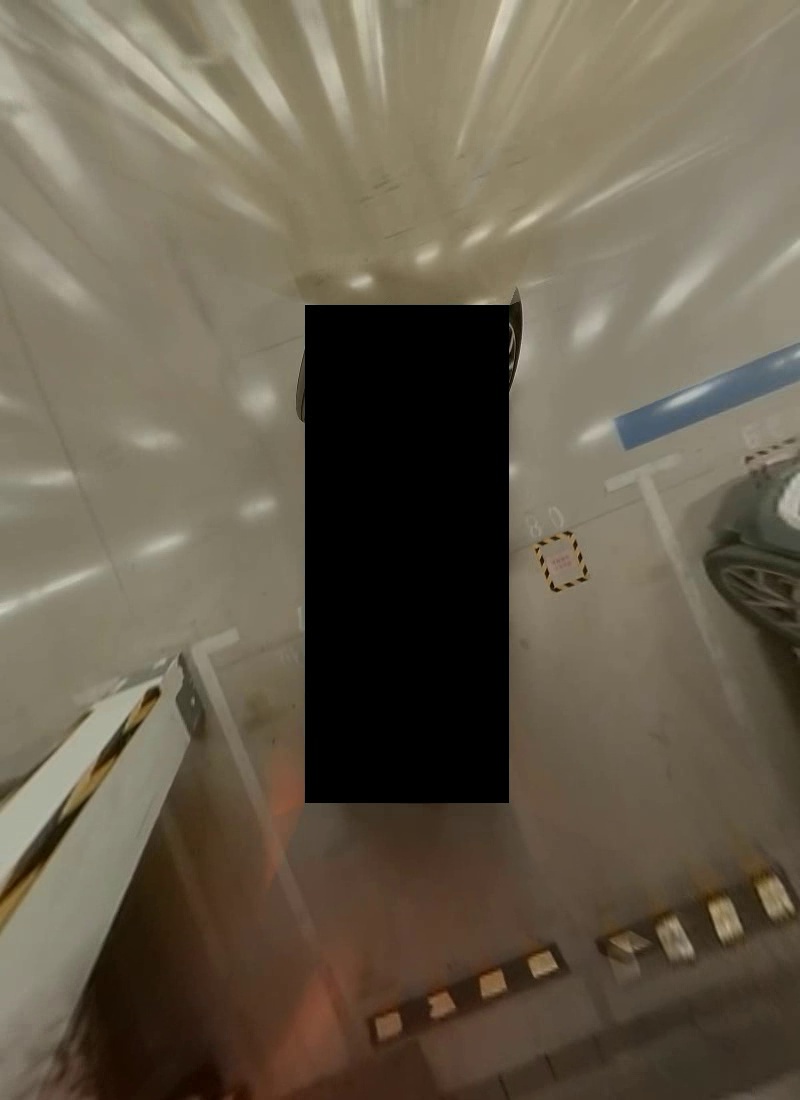}}
        \hfill
        \subfloat[flare with error]{\includegraphics[width=0.31\linewidth]{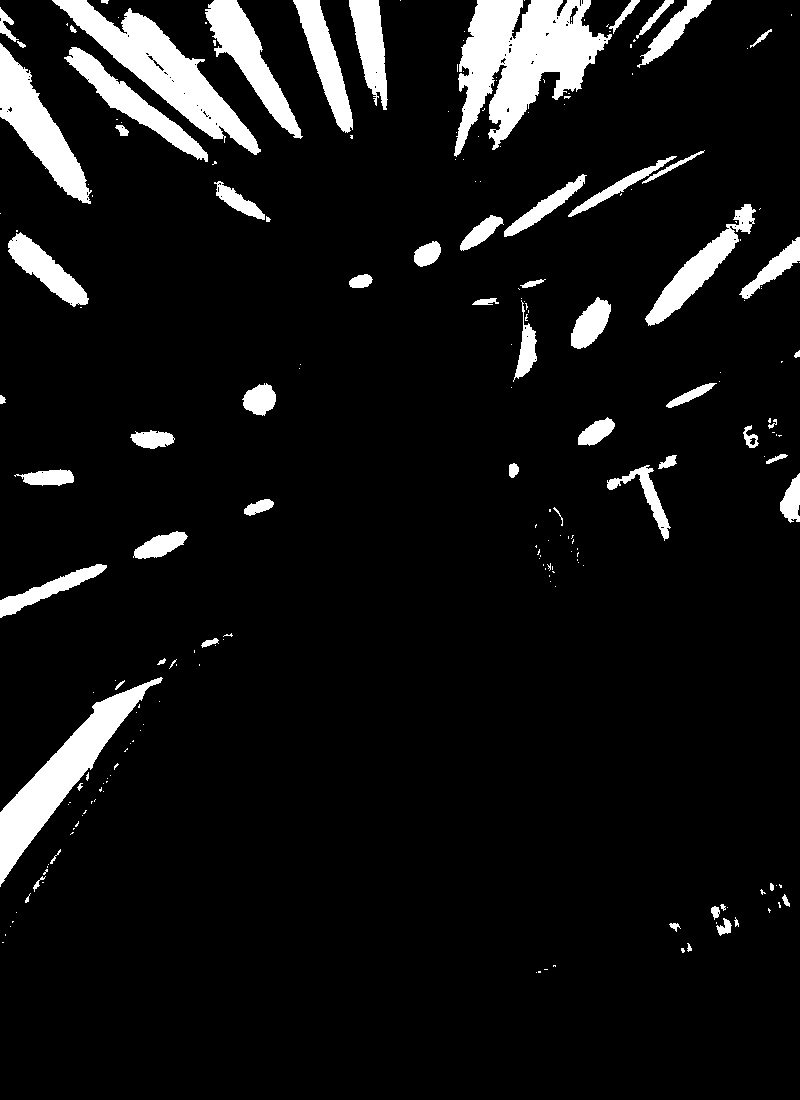}}
        \hfill
        \subfloat[seg. with flare]{\includegraphics[width=0.31\linewidth]{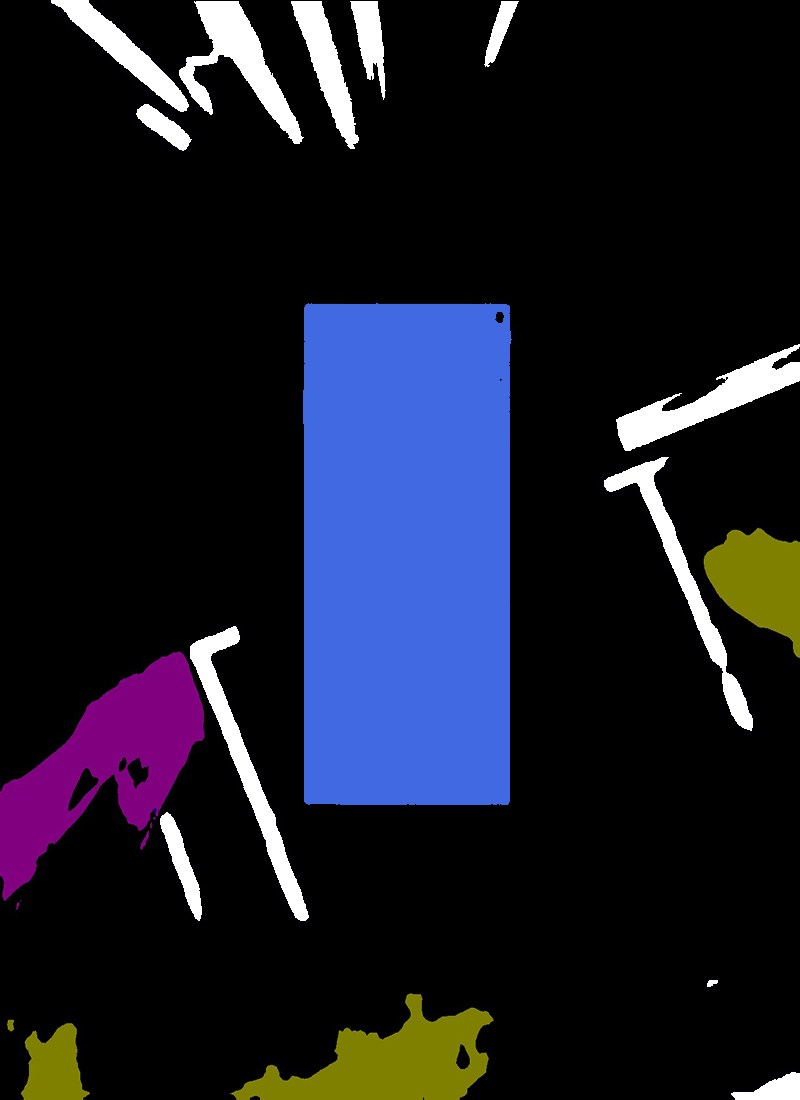}}

        \subfloat[flare removal]{\includegraphics[width=0.31\linewidth]{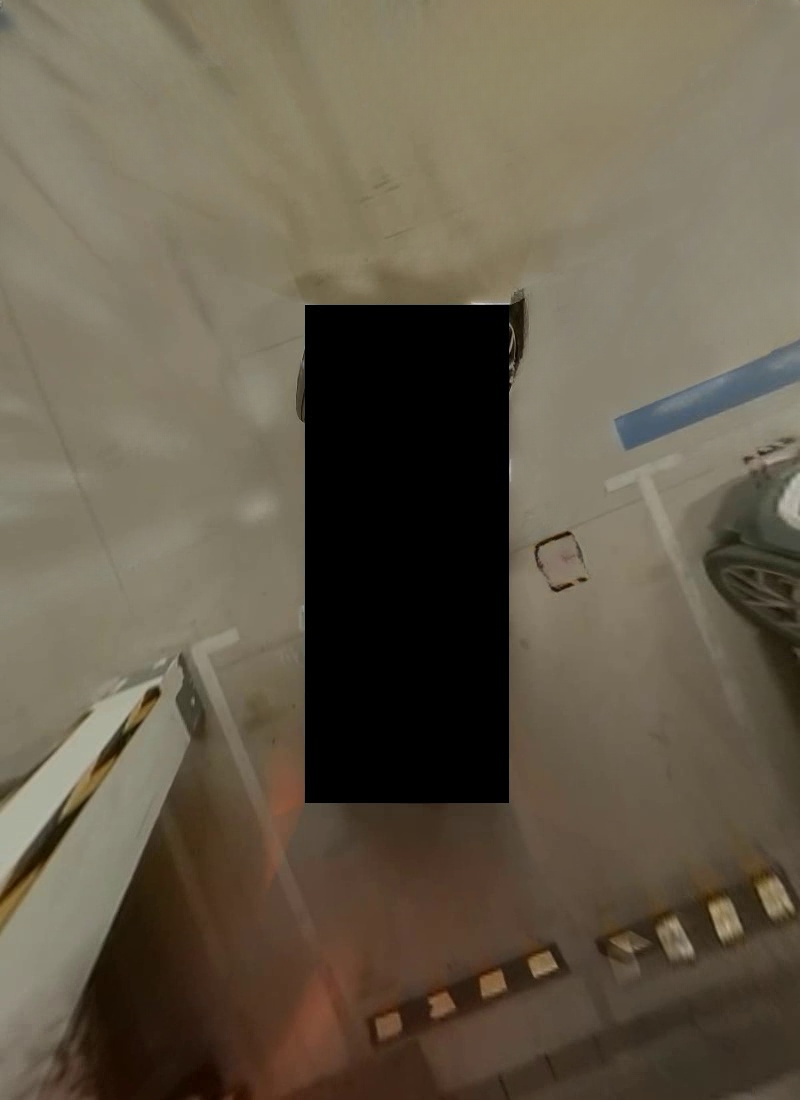}}
        \hfill
        \subfloat[flare without error]{\includegraphics[width=0.31\linewidth]{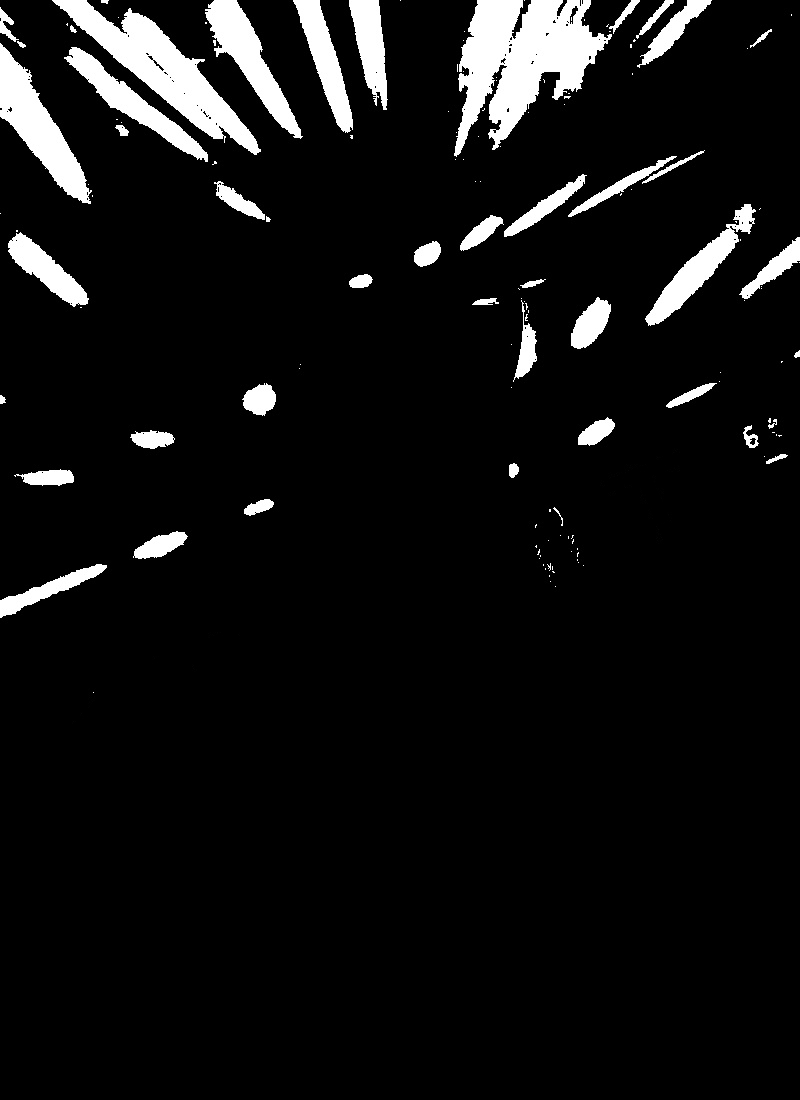}}
        \hfill
        \subfloat[seg. without flare]{\includegraphics[width=0.31\linewidth]{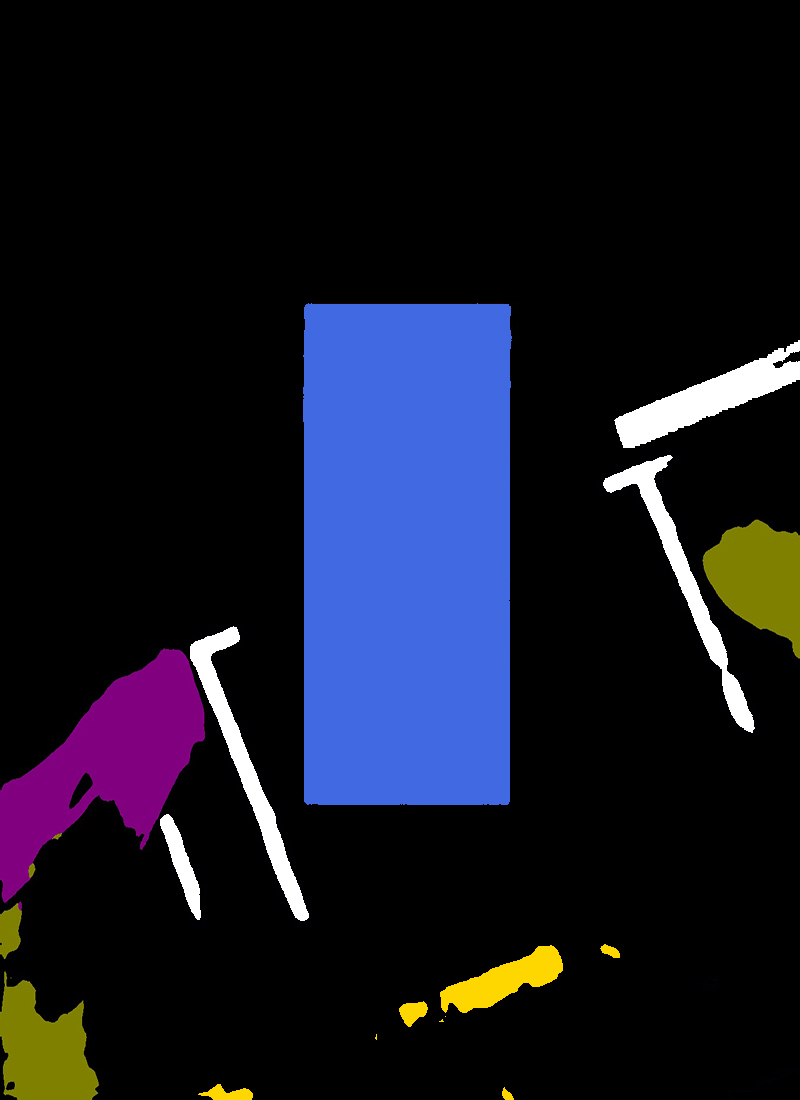}}  
    \caption{\small Flare removal and semantic segmentation.}
    \label{fig_flare}
    \vspace{-0.5em}
\end{figure}

Reflective light on the ground causes diverging flares in the BEV (Fig. \ref{fig_flare}a), interfering with the accurate extraction of road markings (Fig. \ref{fig_flare}c). To tackle this issue, we propose a flare removal strategy based on the U-Net architecture \cite{ronneberger2015u}, aiming at improving semantic segmentation performance.

The process begins with the generation of a highlight mask (Fig. \ref{fig_flare}b) using a specular highlight detection algorithm \cite{li2019specular}. This mask, however, may erroneously include objects such as road markings and king tower as flares. To enhance extraction accuracy, the mask is refined by merging it with manually annotated foreground information, effectively minimizing false positives (Fig. \ref{fig_flare}e). An image inpainting algorithm \cite{suvorov2022resolution} is then applied to remove the identified highlights (Fig. \ref{fig_flare}d), significantly improving segmentation quality (Fig. \ref{fig_flare}f).

The loss function implemented for flare removal is defined as: $$
L = \lambda_{1} \cdot L_1 + \lambda_{2} \cdot L_{\text{p}}, \\
\eqno{(2)}
$$where $L_{\text{p}}$ denotes the perceptual loss \cite{johnson2016perceptual}, and both $\lambda_{1}$ and $\lambda_{2}$ are set to 0.5. By leveraging these two loss functions, we achieve precise flare removal at the pixel level and enhance the visual quality of the images post-removal.

This approach simplifies the labor-intensive process of labeling flare-removed data, thereby streamlining the development of more robust segmentation models.

\begin{itemize}
    \item Semantic Extractor
\end{itemize}

The garage's road markings, including lane lines, parking spots, zebra crossings, and indicating arrows, exhibit enduring stability and remain perspective-invariant. These qualities make them ideal for semantic visual mapping and vehicle localization.

We use DDRnet \cite{hong2021deep} as our semantic feature extraction module, which balances efficiency and accuracy, to segmentation road markings from the BEV images. Fig. \ref{fig_flare}e illustrates the semantic labels, and Fig. \ref{fig_flare}f showcases the segmentation results.

After segmentation, the results are downsampled for efficiency and then reconstructed into a 3D space in the virtual BEV camera coordinate system, as shown below:$$
\left [\begin{array}{c} x_{c} \\ y_{c} \\  z_{c} \end{array} \right ] = K^{-1} * \\
\left [\begin{array}{c} u_{bev} \\ v_{bev} \\ 1 \end{array} \right ], \\
\eqno{(3)}
$$where $K$ is the intrinsic parameter of the virtual BEV camera. $K^{-1}$ is the inverse of the $K$. $\left [\begin{array}{ccc} x_{c} & y_{c} & z_{c} \end{array} \right ]$ is the 3D coordinate value in the virtual BEV camera coordinate system, in this case, the $z_{c}$ of the semantic features on the ground are always equal to 1 meter.

Finally, transfer the 3D semantic features from virtual BEV camera coordinate into vehicle coordinate as follows:$$
\left [\begin{array}{c} x_{v} \\ y_{v} \\ z_{v} \\ 1 \end{array} \right ] = T_{vc} * \\
\left [\begin{array}{c} x_{c} \\ y_{c} \\ z_{c} \\ 1 \end{array} \right ], \\
\eqno{(4)}
$$where $\left [\begin{array}{ccc} x_{v} & y_{v} & z_{v} \end{array} \right ]$ is the 3D coordinate value in the vehicle coordinate system. $T_{vc}$ is the transformation matrix from virtual BEV camera coordinate system to vehicle coordinate system, which is calibrated offline.

\begin{itemize}
    \item Semantic Matcher
\end{itemize}

In this task, we employ the iterative closest point (ICP) algorithm to match 3D semantic features. The cumulative error is mitigated by implementing frame-to-submap matching. This approach enhances efficiency and robustness, particularly when equipped with a reliable initial pose estimation. The initial pose estimation is facilitated by a pose predictor that integrates data from an IMU and a wheel encoder.

\subsection{Pose Predictor}

\begin{itemize}
    \item System Initialization
\end{itemize}

The proposed AVM-SLAM system centers on the BEV semantic features, with other sensors designed to be pluggable. In order to properly initialise the system, we check the availability of data in the data queue of the selected sensors according to the set fusion mode. The system is initialised only if all the selected sensors have data in their data queues. At this point, the vehicle coordinate system is the initial coordinate system for both the global map and the first submap.

\begin{itemize}
    \item Pose Prediction
\end{itemize}

There is evidence that the linear velocity accuracy of the wheel odometry is higher when the vehicle is doing linear motion, and the angular velocity accuracy of the IMU is higher when it is doing rotational motion, and the two are evidently complementary. In order to improve the accuracy and robustness of the pose prediction, this paper adopts the EKF approach to weighted fusion multiple sensors’ data.

Along with the multi-sensor weighted fusion pose prediction, we also pre-integrated the IMU and wheel odometry data between two consecutive keyframes for further optimization of the global pose-graph.

\subsection{SubMapper and GlobalMapper}

To enhance the efficiency of semantic feature ICP matching between frames and the map, we employ a keyframe-submap-globalmap structure for constructing the semantic map. Semantic frames are filtered by the KeyFrameFilter and inserted into the submap if they exhibit more than $50\%$ difference with the previous keyframe. Each submap contains a fixed number of keyframes, typically 10 frames, but this can be adjusted as needed. The number of semantic points in the submap is significantly lower than in the global map, improving efficiency during frame-to-submap semantic feature ICP matching and reducing error accumulation in frame-to-frame matching.

Within the Mapping module, we maintain two submaps: the current submap and the upcoming submap, ensuring sufficient co-visibility area between adjacent submaps (set at $50\%$ in this system). Keyframes are simultaneously inserted into both submaps. Once the maximum number of keyframes is reached in the current submap, we execute point cloud correction and local optimization. Then, the current submap is integrated into the global map, and the next submap takes its place, starting the creation of a new subsequent submap.

\subsection{Loop Detector}

Loop detection serves as a pivotal constraint for enabling global optimization and significantly influences both mapping scale and speed. In order to make the map building results better, loop points need to be added to perform global optimisation. However, there are numerous repetitive textures and similar structures in the underground garage, so not all keyframes and submaps are suitable for loop detection. To address this, we have devised an effective mechanism known as SPQ to filter potential loop frames and submaps, thus reducing the number of detection and preventing mismatches.

The mechanism evaluates the feasibility of candidate loop frames and submaps based on the categories of semantic features and their weights in the keyframes and submaps. Scores are assigned to these candidate frames and submaps, with higher scores for more categories and higher weights for rarer features. When the score exceeds a pre-determined threshold, the keyframes and submaps are eligible to become candidate loop frames and submaps for subsequent loop detection. The SPQ-LoopDetection method is outlined in Algorithm \ref{alg_SPQ} for reference.

\begin{algorithm}[h]
\caption{SPQ-LoopDetection method}
\label{alg_SPQ}
\begin{algorithmic}[1]
\Require{KeyFrame(kf)}
    \State $PreScore = FeatureItem(kf)$
    \If{PreScore $>$ ThreshPreScore}
        \For{map in LoopMaps}
            \State $distance = {|kf.pose - map.pose|}$
            \If{distance $<$ ThreshDistance}
                \State $score, transform = Match(kf, map)$
                \If{score $>$ ThreshMatchScore}
                    \State $loopResult += kf, map, transform$
                \EndIf
            \EndIf
        \EndFor
    \EndIf
\State $CurrentMap += kf$
\If{CurrentMap.finish}
    \State $PreScore = FeatureItem(CurrentMap)$
    \If{PreScore $>$ ThreshPreScore}
        \State $LoopMaps += CurrentMap$
    \EndIf
\EndIf
\end{algorithmic}
\end{algorithm}

\subsection{Global Optimizer}

\begin{figure}[t]
    \captionsetup{font=small,labelfont=bf}
    \centering
    \includegraphics[width=0.95\linewidth]{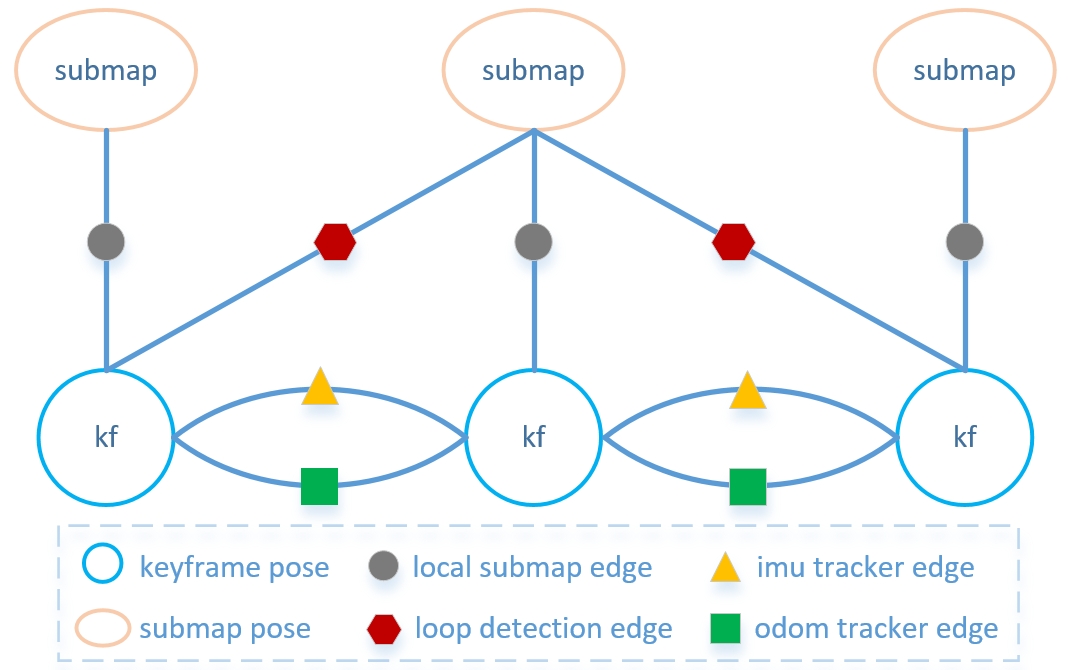}
    \caption{\small Schematically of pose-graph with additional kinematic constraints.}
    \label{fig_optimize}
    \vspace{-1.0em}
\end{figure}

We employ a pose-graph approach for global optimization. As depicted in Fig. \ref{fig_optimize}, the pose-graph's Nodes encompass both keyframes and submaps, while Edges represent constraints involving keyframe-to-keyframe and keyframe-to-submap. Keyframe-to-keyframe constraints encompass semantic visual constraints between adjacent keyframes, alongside additional kinematic constraints derived from pre-integrated values (IMU and Wheel). Keyframe-to-submap constraints involve semantic visual constraints between keyframes and submaps, along with semantic loop constraints obtained from semantic loop detection. The global optimizer periodically performs optimization operations on the collected Nodes and Edges, subsequently updating the results for each keyframe and submap.

\section{EXPERIMENTS}

\subsection{Benchmark Dataset}

To validate the proposed AVM-SLAM system, tests were conducted in a $220m\times110m$ underground garage with over 430 parking spots using a test vehicle equipped with four surround-view fisheye cameras, four wheel encoders, and an IMU, all synchronized and calibrated offline. The four fisheye cameras formed an AVM subsystem for real-time BEV stitching at 30Hz.

The proposed benchmark dataset to be publicly included four fisheye image sequences, one BEV image sequence, four wheel encoder data, and one IMU data. Fisheye images had a resolution of $1280\times960$, and BEV images had a resolution of $1354\times1632$, representing a physical area of $14.25m\times17.18m$ (1.05cm per pixel on the ground). Both image sequences were stored at 10Hz. Our system used the BEV image as input for validation, while other visual algorithms used front fisheye camera images, which did not affect their operation.

\subsection{Ablation Study}

\begin{figure*}[th]
    \captionsetup{font=small,labelfont=bf}
    \centering
        \subfloat[VIWFusion]{\includegraphics[width=0.24\linewidth, height=1.5cm]{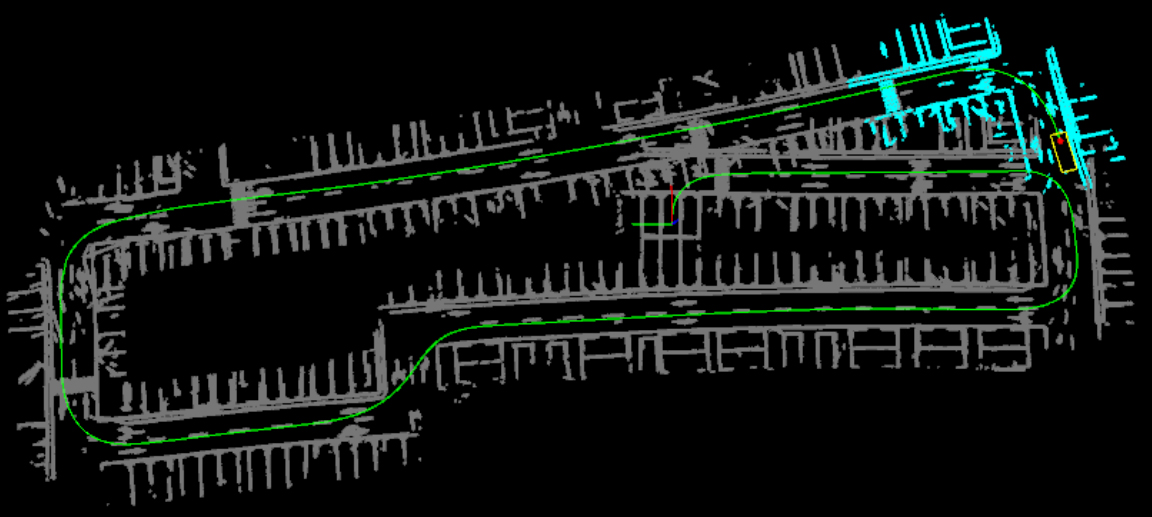}}
        \hfill
        \subfloat[Ordinary (sim. AVP-SLAM)]{\includegraphics[width=0.24\linewidth, height=1.5cm]{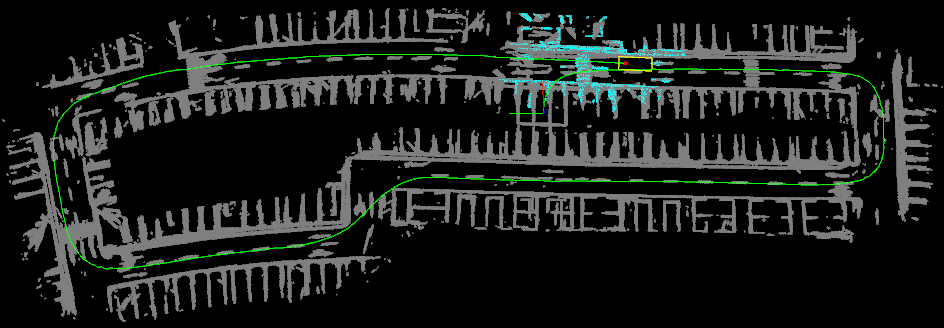}}
        \hfill
        \subfloat[SPQ+Additional]{\includegraphics[width=0.24\linewidth, height=1.5cm]{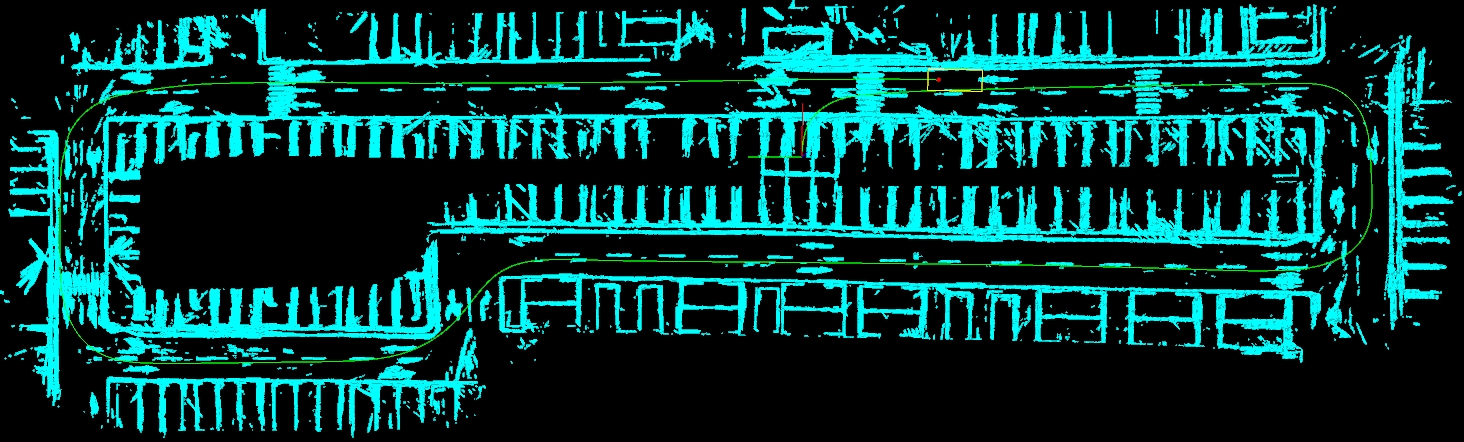}}
        \hfill
        \subfloat[SPQ+Additional+FlareRemoval]{\includegraphics[width=0.24\linewidth, height=1.5cm]{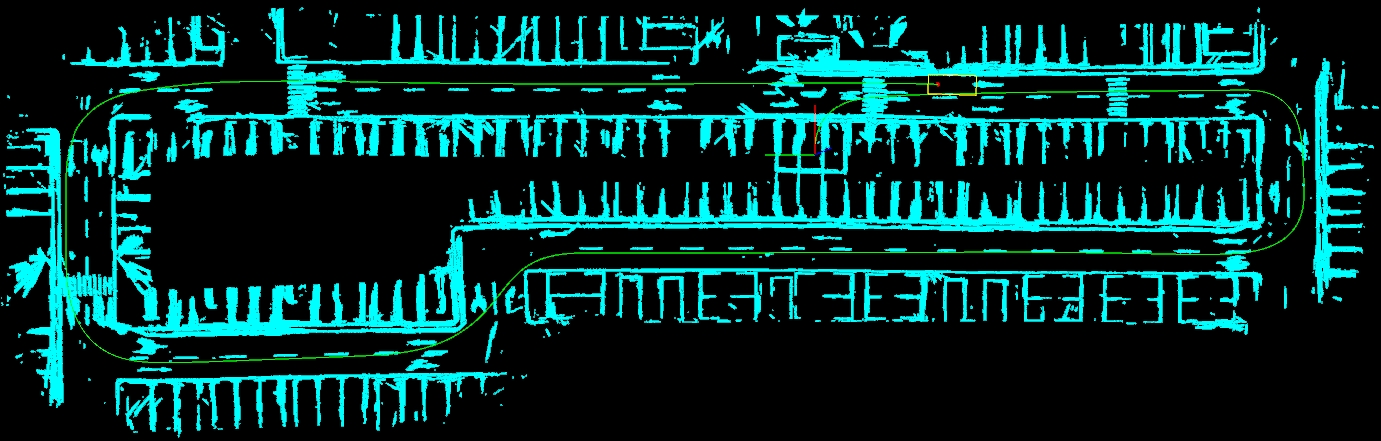}}

        \subfloat[VIWFusion]{\includegraphics[width=0.24\linewidth, height=2.2cm]{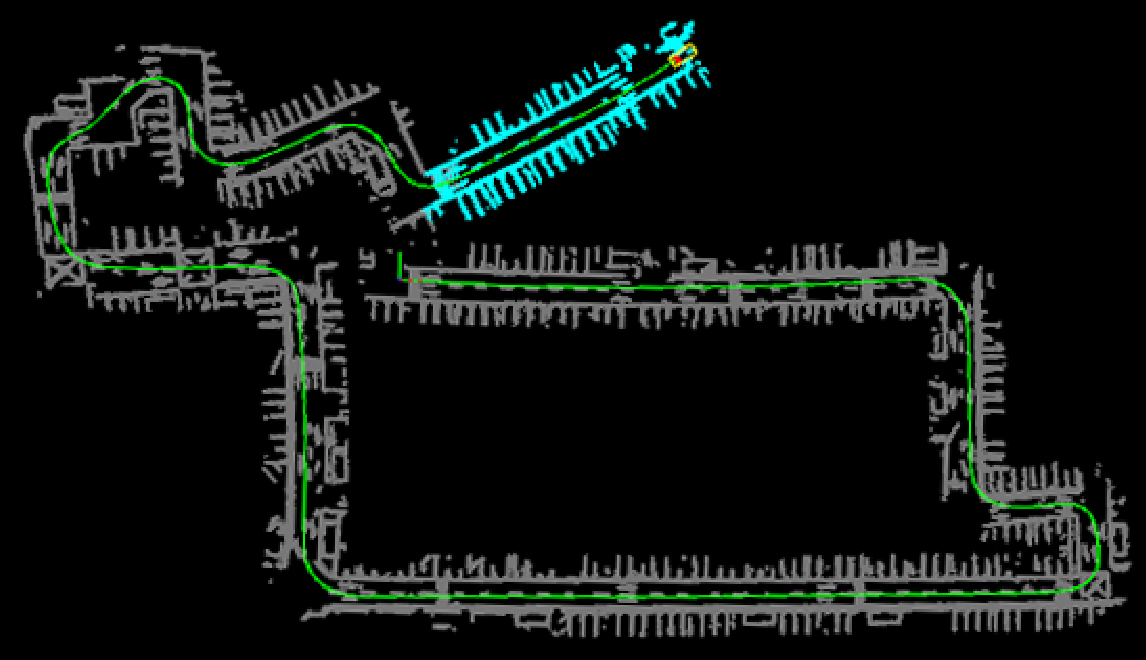}}
        \hfill
        \subfloat[Ordinary (sim. AVP-SLAM)]{\includegraphics[width=0.24\linewidth, height=2.2cm]{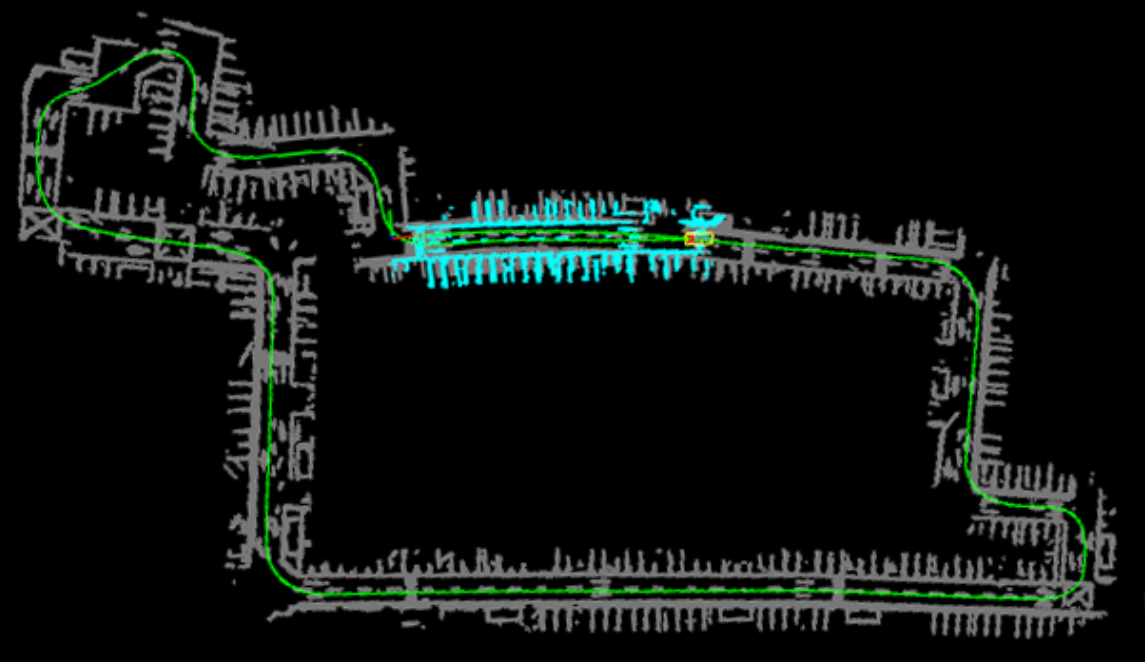}}
        \hfill
        \subfloat[SPQ+Additional]{\includegraphics[width=0.24\linewidth, height=2.2cm]{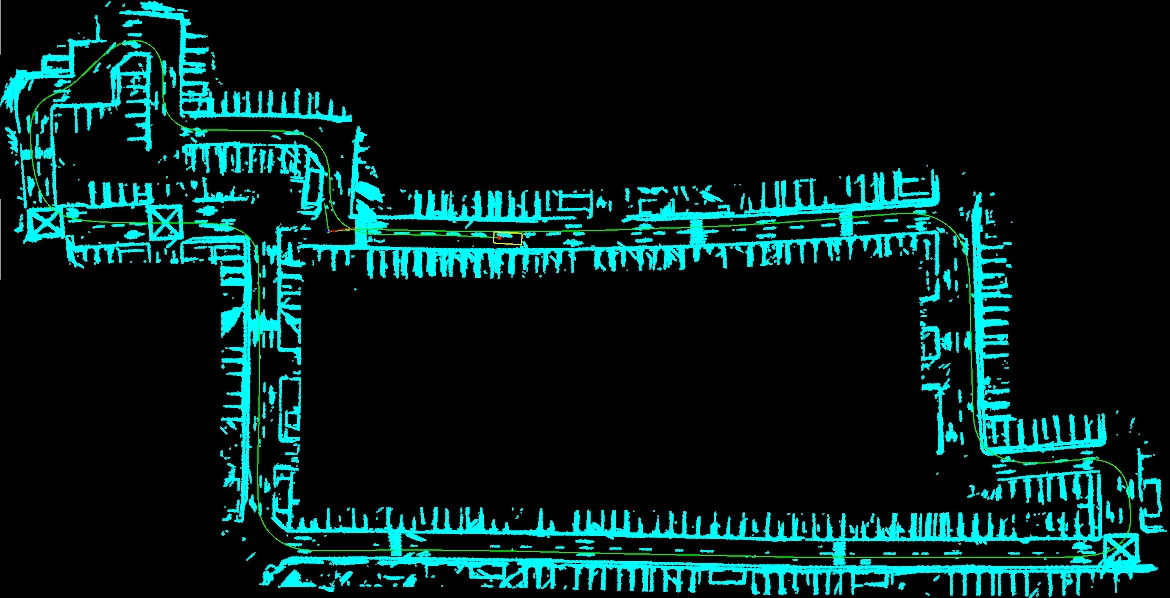}}
        \hfill
        \subfloat[SPQ+Additional+FlareRemoval]{\includegraphics[width=0.24\linewidth, height=2.2cm]{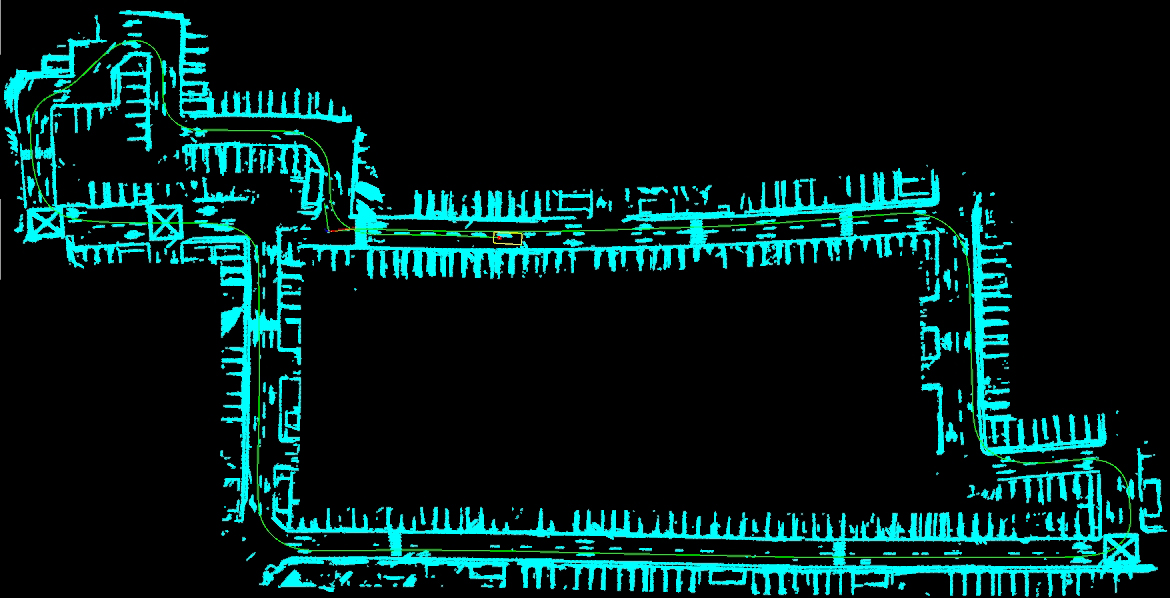}}

        \subfloat[VIWFusion]{\includegraphics[width=0.24\linewidth, height=2.2cm]{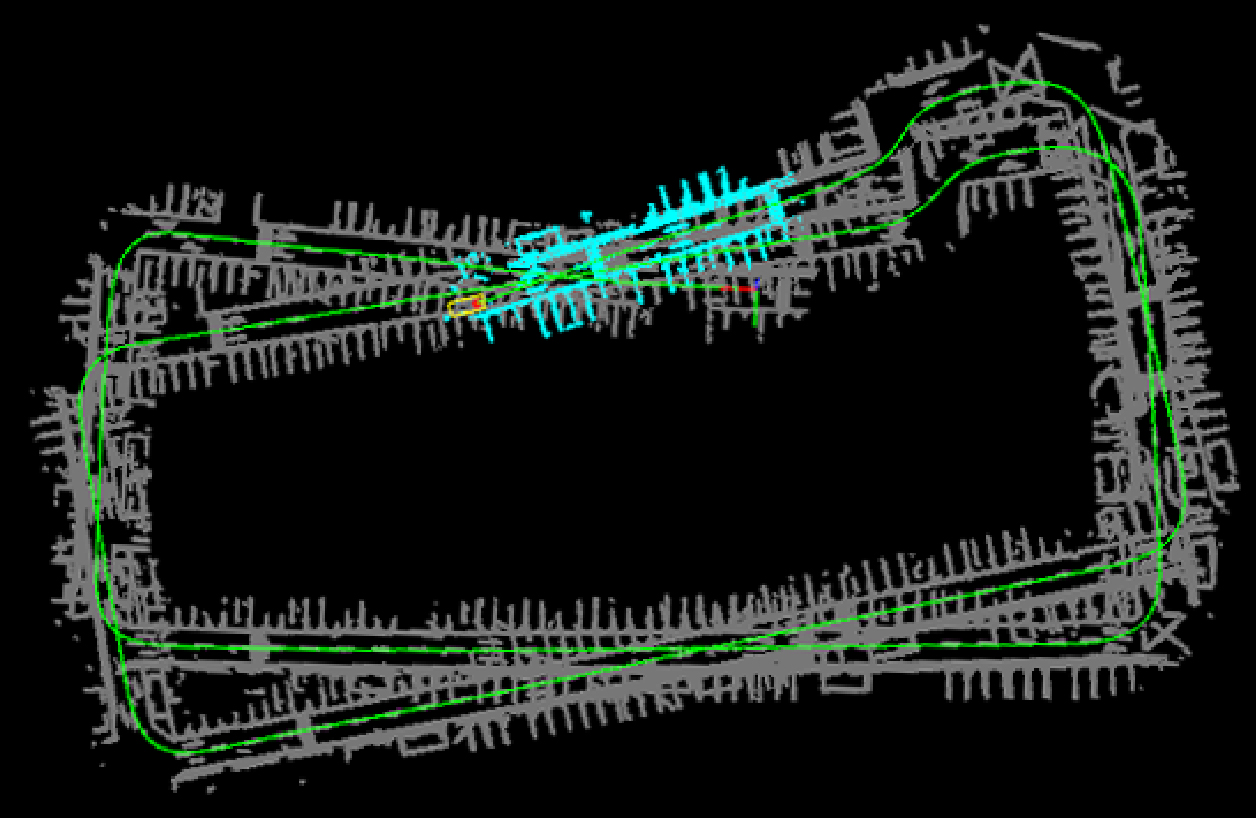}}
        \hfill
        \subfloat[Ordinary (sim. AVP-SLAM)]{\includegraphics[width=0.24\linewidth, height=2.2cm]{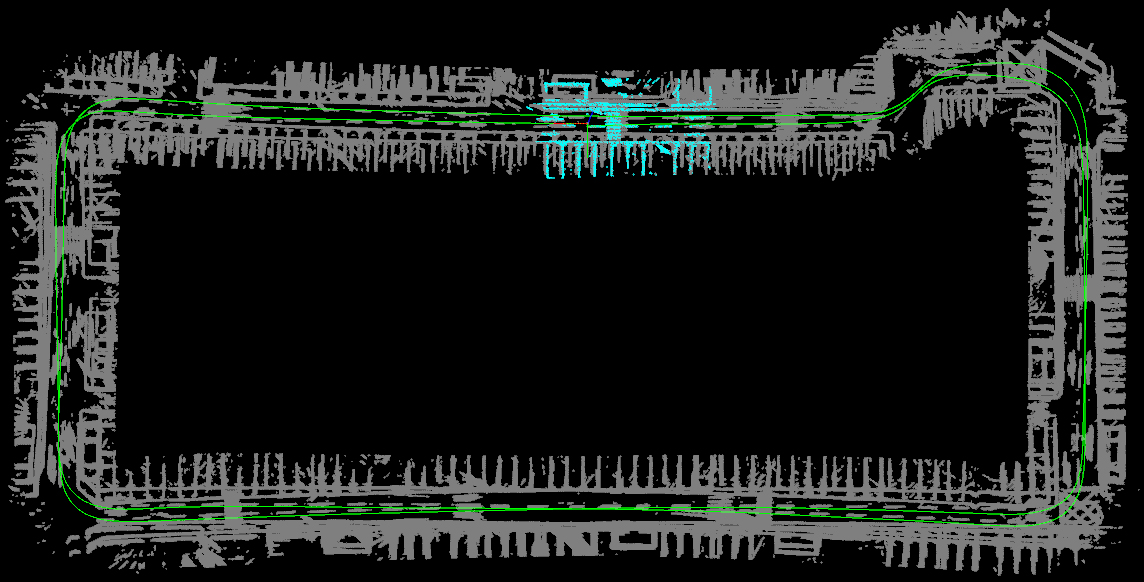}}
        \hfill
        \subfloat[SPQ+Additional]{\includegraphics[width=0.24\linewidth, height=2.2cm]{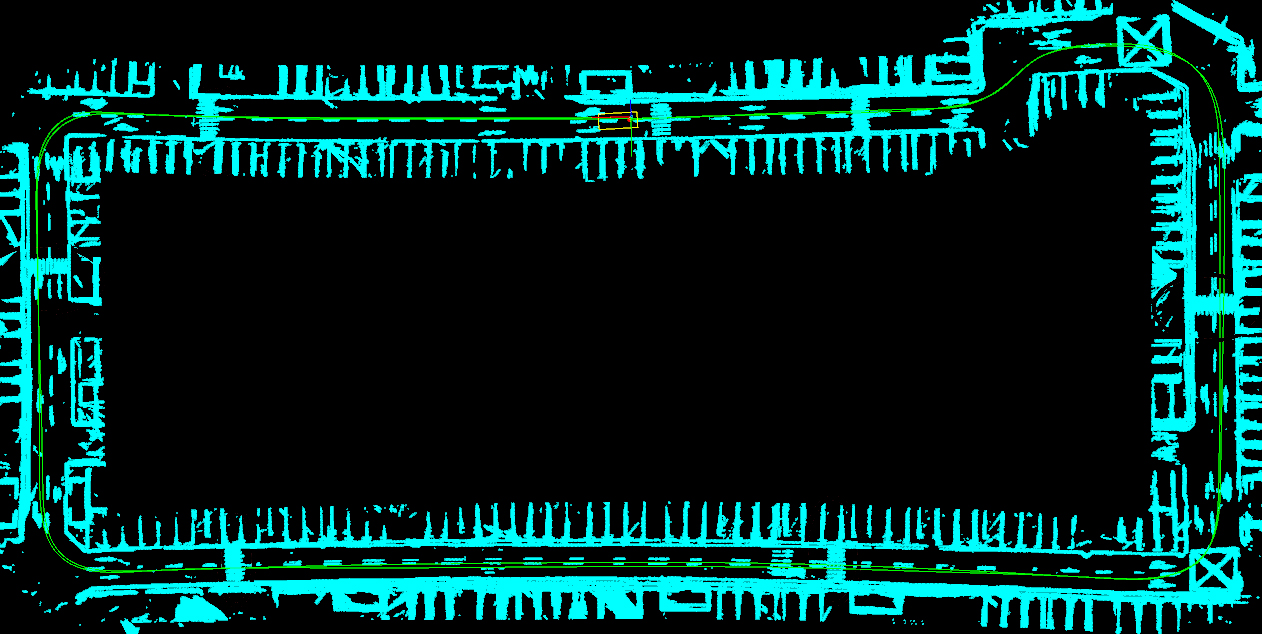}}
        \hfill
        \subfloat[SPQ+Additional+FlareRemoval]{\includegraphics[width=0.24\linewidth, height=2.2cm]{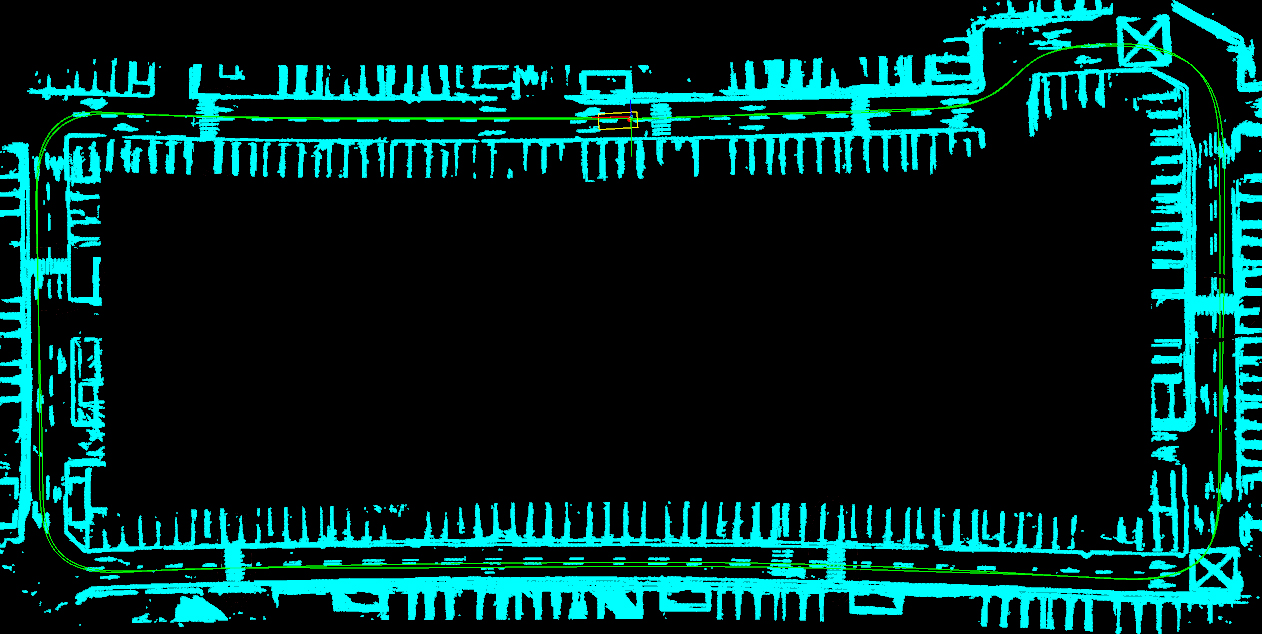}}
        
    \caption{\small Mapping results by VIWFusion, ordinary loop detection, SPQ loop detection \& additional kinematic constraints, and flare removal.}
    \label{fig_mapping}
    \vspace{-1.0em}
    
\end{figure*}

We compared the semantic segmentation results (Fig. \ref{fig_flare} and Tab. \ref{tab_flare}) and mapping results (last two columns of Fig. \ref{fig_mapping} and Tab. \ref{tab_error}) before and after flare removal. From the results, we can find that the flare removal module substantially improves the accuracy of semantic segmentation and reduces the flares introduced into the map due to mis-segmentation, thereby obtaining a more perfect map.

\begin{table}[h]
    \setlength{\belowcaptionskip}{0.5em}
    \setlength{\abovecaptionskip}{0em}
    \captionsetup{font=small,labelfont=bf}
    \caption{\small Effect of flare removal on semantic segmentation results.}
    \label{tab_flare}
    \begin{center}
    \begin{tabular}{c|ccc}
        \toprule
        \textbf{Methods} & \textbf{Model Size} & \textbf{FPS} & \textbf{MIoU} \\
        \midrule
        \multirow{2}{*}{Original BEV} & 48MB & 107 & 0.58 \\
                                      & 23MB & 166 & 0.54 \\
         \midrule
         \multirow{2}{*}{Flare Removal} & 48MB & 107 & 0.68 \\
                                        & 23MB & 166 & 0.67 \\
        \bottomrule
    \end{tabular}
    \end{center}
\end{table}

We also performed ablation experiments on the SPQ module (Fig. \ref{fig_mapping} and Tab. \ref{tab_error}). It is clear that there is an unavoidable cumulative error in the VIWFusion front-end (first column of Fig. \ref{fig_mapping}). And the global optimisation using ordinary loop detection (second column of Fig. \ref{fig_mapping}) is prone to fail in complex environments such as garages where there are numerous repetitive textures and similar structures. On the contrary, the global optimisation with SPQ loop detection and additional kinematic constraints (third column of Fig. \ref{fig_mapping} and Tab. \ref{tab_error}) shows better mapping results.

\subsection{Robustness and Accuracy of Mapping}

\begin{itemize}
    \item Robustness
\end{itemize}
    
We used images from a front-mounted fisheye camera as input and tried to run feature-based ORB-SLAM3 [3], optical flow based SVO [4] and DSO [5], and vision-inertia fusion based VINS-Mono [9]. Unexpectedly the traditional visual SLAM for the above state-of-the-art (SOTA) all suffer from initialization failures, frequent loss of tracking, and runtime failure because of the poor lighting, sparse texture, and scene variability of garage. On the other hand, the method in this paper is very stable because it adopts the vision-inertia-wheel fusion method for pose tracking and uses the semantic features of the road marking extracted from the BEV to construct the map. The tracking and mapping are stable using datasets under different conditions, which confirms the robustness and reliability of our algorithm.

\begin{itemize}
    \item Accuracy
\end{itemize}

\begin{figure}[t]
    \captionsetup{font=small,labelfont=bf}
    \vspace{-4pt}
    \centering
        \subfloat[schematic map of garage]{\includegraphics[width=0.485\linewidth]{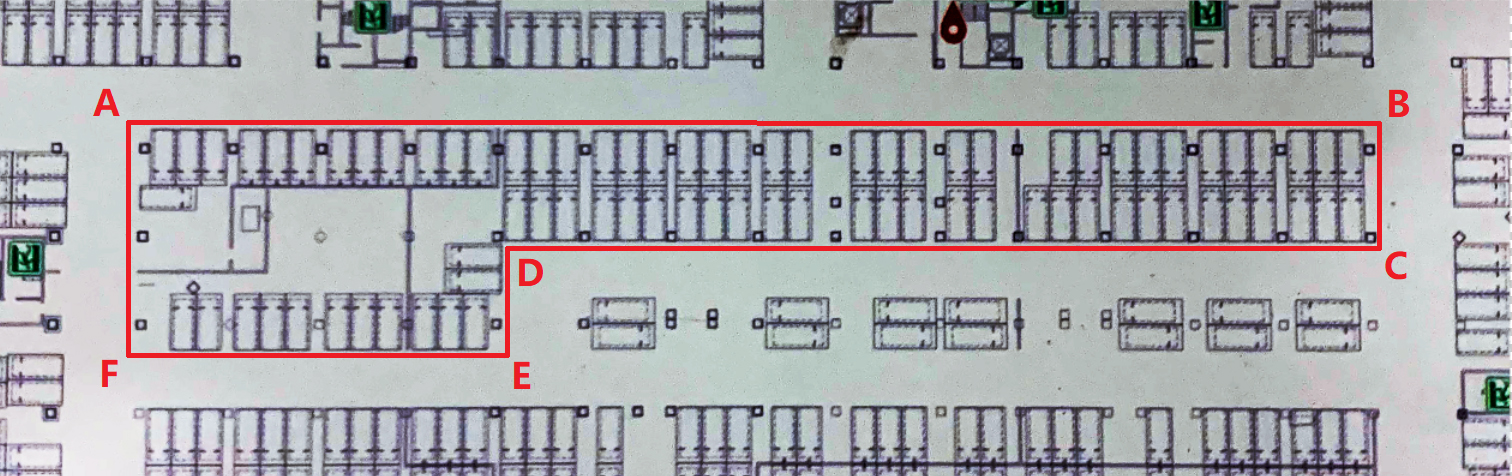}}
        \hfill
        \subfloat[overlapping comparison]{\includegraphics[width=0.485\linewidth]{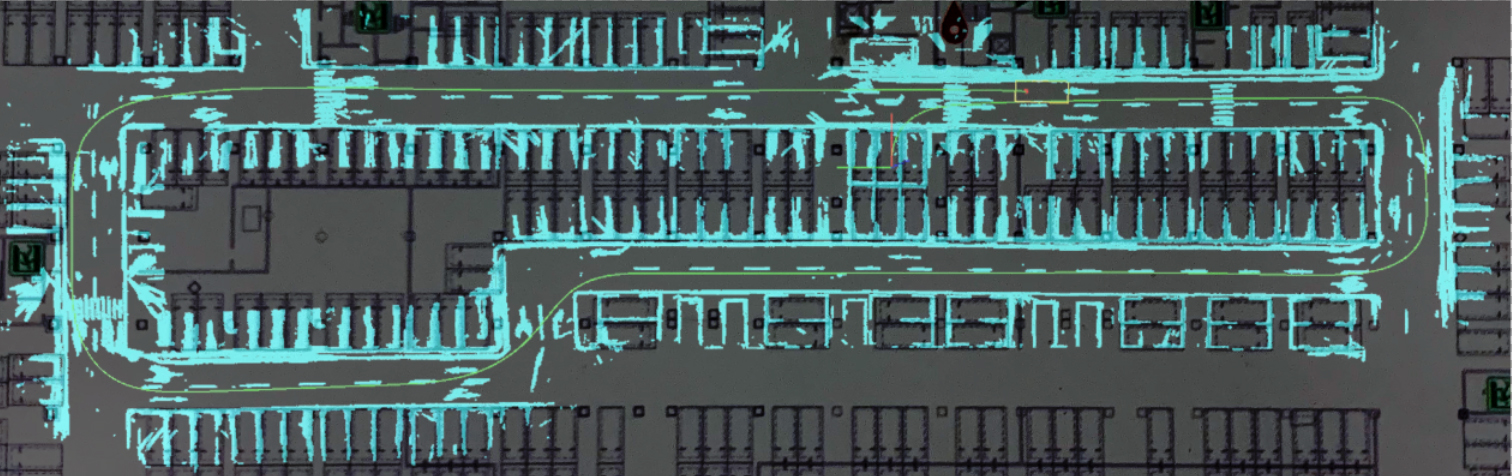}}
    \caption{\small Qualitatively results of overlapping comparison.}
    \label{fig_overlapping}
    \vspace{-1.0em}
\end{figure}

Firstly, we performed a qualitative analysis using a schematic map of the planar structure of the garage (as shown in Fig. \ref{fig_overlapping}a) with the semantic map we constructed. The two are made to have the same size by scaling and are overlapped for comparison (as shown in Fig. \ref{fig_overlapping}b). Obviously, the semantic map built by our system can be perfectly aligned with the schematic map of the garage, with very high mapping accuracy.

Secondly, we performed a quantitative analysis comparing the world distances between selected points in multiple datasets (denoted as ABCDEF in dataset 1, as shown in Fig. \ref{fig_overlapping}a) and the corresponding map distances. The world distances were measured using a high-precision laser rangefinder, while the map distances were computed from the respective 3D point coordinates.

Regrettably, to date, we have not obtained the specific implementation of the AVP-SLAM \cite{qin2020avp}, which is the most closely related superior work to ours. However, it is not difficult to observe that the ordinary implementation of our paper, with the flare removal module and SPQ module removed, can be used to simulate AVP-SLAM \cite{qin2020avp}. The experimental results (Tab. \ref{tab_error}) confirm that, among all the test data, ours AVM-SLAM system augmented with SPQ loop detection, additional kinematic constraints, and flare removal module achieves the highest mapping accuracy.

\begin{table}[h]
    \setlength{\belowcaptionskip}{0.5em}
    \setlength{\abovecaptionskip}{0em}
    \captionsetup{font=small,labelfont=bf}
    \caption{\small Absolute Error of mapping by diffrent modules. The ordinary implementation can be used to simulate AVP-SLAM \cite{qin2020avp}.}
    \label{tab_error}
    \begin{center}
    \begin{tabular}{c|c|ccc}
        \toprule
        \textbf{Dataset} & \textbf{Modules} & \textbf{MAE [m]} & \textbf{Max [m]} & \textbf{RMSE [m]} \\
        \midrule
        \multirow{3}{*}{data1} & Ordinary & 2.845 & 9.048 & 4.122 \\
                               & SPQ+Add. & 0.791 & 1.807 & 1.047 \\
                               & SPQ+Add.+Flare. & \textbf{0.651} & \textbf{1.413} & \textbf{0.785} \\
        \midrule
        \multirow{3}{*}{data2} & Ordinary & 3.032 & 7.740 & 3.906 \\
                               & SPQ+Add. & 1.814 & \textbf{2.502} & 2.008 \\
                               & SPQ+Add.+Flare. & \textbf{1.632} & 2.537 & \textbf{1.903} \\
        \midrule
        \multirow{3}{*}{data3} & Ordinary & 1.747 & 4.479 & 2.234 \\
                               & SPQ+Add. & \textbf{0.871} & 1.882 & 1.139 \\
                               & SPQ+Add.+Flare. & 0.936 & \textbf{1.725} & \textbf{1.067} \\
        \bottomrule
    \end{tabular}
    \end{center}
    \vspace{-1.0em}
\end{table}

\section{CONCLUSIONS}

In this paper, we introduce the AVM-SLAM system for AVP tasks in the indoor parking lots. A distinguishing feature of our system is the implementation of flare removal technology within BEV images for the first time, significantly enhancing road marking detection capabilities and achieving excellent mapping and localization performance. Additionally, we have pioneered a SPQ module that effectively addresses the challenges presented by environments with repetitive textures, thereby improving loop detection and system robustness. Furthermore, we provide a multi-sensor, high-resolution benchmark dataset for garage localization and mapping development and evaluation. Experimental results validate our approach for AVP tasks in garages. Future work includes refining the AVM-SLAM framework, optimizing multi-sensor fusion, and improving flare removal and semantic segmentation models.


\section*{ACKNOWLEDGMENT}

This research was partly supported by the National Key R\&D Program of China (No.2023YFB3308601), Science and Technology Service Network Initiative (No. KFJ-STS-QYZD-2021-21-001), the Talents by Sichuan provincial Party Committee Organization Department, and Chengdu - Chinese Academy of Sciences Science and Technology Cooperation Fund Project (Major Scientific and Technological Innovation Projects).

\bibliographystyle{ieeetr}
\bibliography{ref}

\end{document}